\documentclass{article}

\usepackage[table]{xcolor}

\usepackage{color}
\usepackage{tikz}
\usepackage{tikz-qtree}

\usepackage[nonatbib,final]{nips_2017}

\usepackage[utf8]{inputenc} 
\usepackage[T1]{fontenc}    
\usepackage{hyperref}       
\usepackage{url}            
\usepackage{booktabs}       
\usepackage{amsfonts}       
\usepackage{nicefrac}       
\usepackage{microtype}      
\usepackage{amsmath}        

\title{Globally Optimal Symbolic Regression}

\author{
  Vernon Austel\thanks{IBM T.J. Watson Research Center,  Yorktown Heights, NY 10598.\newline \texttt{\{austel,sanjeebd,gunluk,lhoresh,nannicini,sbar\}@us.ibm.com}}
\And Sanjeeb Dash\footnotemark[1]
\And Oktay Gunluk\footnotemark[1]
\And Lior Horesh\footnotemark[1]\\
\And Leo Liberti\thanks{Ecole Polytechnique, 91128 Palaiseau, Paris, France.  \texttt{liberti@lix.polytechnique.fr}}
\And Giacomo Nannicini\footnotemark[1]
\And Baruch Schieber\footnotemark[1]}

\author{
	Vernon Austel\thanks{IBM T.J. Watson Research Center,  Yorktown Heights, NY 10598.\newline .~~~~~~~~~\texttt{\{austel,sanjeebd,gunluk,lhoresh,nannicini,sbar\}@us.ibm.com}}
	~~ Sanjeeb Dash\footnotemark[1]
	~~ Oktay Gunluk\footnotemark[1]
	~~ Lior Horesh\footnotemark[1]\\[.2cm]
	\bf Leo Liberti\thanks{Ecole Polytechnique, 91128 Palaiseau, Paris, France.  \texttt{liberti@lix.polytechnique.fr}}
	~~ Giacomo Nannicini\footnotemark[1]
	~~ Baruch Schieber\footnotemark[1]}

\begin{document}

\maketitle

\begin{abstract}
  In this study we introduce a new technique for symbolic regression that guarantees global optimality.
This is achieved by formulating a mixed integer non-linear program (MINLP) whose solution is a symbolic mathematical expression of minimum complexity that explains the observations.
We demonstrate our approach by  rediscovering Kepler's law on planetary motion using exoplanet data
  and Galileo's pendulum periodicity equation using experimental data.
\end{abstract}
\section{Introduction}
Discovering mathematical models that explain the behavior of a system
has a broad range of applications.  Symbolic regression is an important
field in machine learning whose goal is to find a symbolic
mathematical expression that explains a dependent variable in terms of
a number of independent variables for a given data set, most commonly
as an explicit function of the independent variables. Unlike
traditional (numerical) regression schemes, the functional form of the
expression is not assumed to be known {\em a priori}
\cite{kroll2017workflow}.  The utility of the approach has been
established for a broad range of applications
\cite{connor1977scaling,willis1997genetic,davidson1999method,davidson2003symbolic},
including the discovery of equations
\cite{todorovski1997declarative,langley1981data,schmidt2009distil,schmidt2010symbolic}.

Starting with \cite{koza}, symbolic regression problems have been typically  solved with genetic programming \cite{banzhaf1998genetic,augusto2000symbolic}, an evolutionary {\it metaheuristic} related to genetic algorithms; another such technique is grammatical evolution \cite{ge}.
Even before \cite{koza}, heuristics to find explicit functional relationships were developed as part of the BACON system, see \cite{langley1987}.
There has been much research into improving symbolic regression techniques.
In \cite{Luo2017}, the separability of the desired functional form is exploited to speed up symbolic regression, whereas a set of candidate basis functions is used in \cite{ffx} for this purpose.
Other recent work has focused on finding accurate constants in the derived symbolic mathematical expression \cite{dgp}.
To discover meaningful functions in the context of physical systems,
the authors of \cite{schmidt2009distil} search for functions that not only match the data, but also have the property that partial derivatives also match the empirically computed partial derivatives. The same authors  search for implicit functional relationships in \cite{schmidt2010symbolic}.
Another approach populates a large hypothesis space of functional forms, on which sparse selection is applied \cite{brunton2016discovering}. 


In this study we present a Mixed-Integer Non-Linear Programming (MINLP) formulation that  produces the simplest symbolic mathematical expression 
 that satisfies a prescribed upper bound on the prediction error. 
Our MINLP formulation can be solved to optimality using existing state-of-the-art global optimization solvers.
The key advantage of our approach is that it produces a {\it globally optimal} mathematical expression while avoiding exhaustive search of the solution space.
Another advantage is that it produces correct real-valued constants (within a tolerance) directly; most other methods use specialized algorithms to refine constants \cite{dgp, topchy} and cannot guarantee global optimality.
In addition, our formulation can, in principle, seamlessly incorporate additional constraints and objectives to capture domain knowledge and user preferences.

\section{A MINLP Formulation for Free-Form Model Discovery}
A  symbolic regression scheme consists of a  space of valid mathematical expressions together with a  mechanism for  its exploration.
A  mathematical expression can be  represented by a (rooted) expression tree where each node is labeled by  one of the following entities: operators (such as $+, -$, $\times$, $\log$), variables (i.e., the independent variables), and constants.
Edges of the tree link these entities in a way that is consistent with a prescribed grammar.
%
%
For example, an expression tree for $2(w_1+w_2)^3+1$ is:
    \begin{center}
	{\small
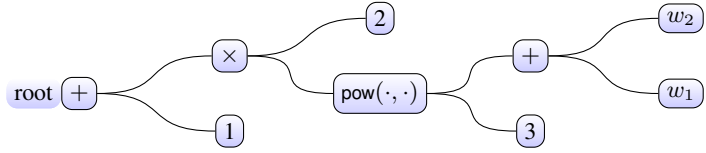
\begin{tikzpicture}[scale=1, sibling distance=1cm,  grow=right, level distance = 2cm,
		edge from parent path=	{(\tikzparentnode.east) .. controls +(1,0) and +(-1,0)	.. (\tikzchildnode.west)}, 
		every node/.style = {shape=rectangle, rounded corners,	draw, align=center,	top color=white, bottom color=blue!20}]]
\node[label=left:{root}] {$+$}
child { node {1} }
child { node {$\times$}
	child { node {$\mbox{\scriptsize\textsf{pow}}(\cdot,\cdot)$}
		child { node {3} }
		child { node {$+$}child { node {$w_1$} }
			child { node {$w_2$} } }
			}
	child { node {2} } };
\end{tikzpicture}
    }\end{center}
Here $+$, $\times$ and the power function are the operators,  $w_1,w_2$ are the variables, and numbers $1$, $2$ and $3$ are the constants. 

In our MINLP based approach we model the grammar of valid expression trees by a set of constraints.  
The discrete variables of the formulation are used to define the structure of the expression tree, and the continuous variables to evaluate the resulting symbolic expression for specific numerical values associated with the data.
To choose among the many
possible expressions that fit the data (within an error bound), we set
the objective to minimize the description complexity of the
expression.  Consequently, the MINLP formulation has the form
\begin{align}
\min&&\mathcal{C}(f_{cqwz} )&									\tag{Complexity}\label{eq:C}\\
\text{s.t.}&&{(c,q,w,z)}			&\in \mathcal{T}   					\tag{Grammar}\label{eq:G}\\
	&&v_i					&=f_{cqwz}(x^i)  ,~~\forall i\in I 	\tag{Prediction}\label{eq:M}\\
	&& \mathcal{D}(v,y) 	&\le \epsilon  						\tag{Error}\label{eq:E}
\end{align}
where $c,q,w$ and $z$ are decision variables, $f_{cqwz}$ is the
expression tree defined by these variables, $\mathcal{T}$ is the set
of values of the decision variables that define valid expression trees
of bounded size, $\mathcal{C}$ measures the description complexity of
the expression tree, $\mathcal{D}$ measures error of the
predicted values $v$, and $(x,y)$ are the observed data.  In
practice, MINLP solvers (e.g., BARON \cite{ts05},
COUENNE\cite{BeLeLiMaWa08}, SCIP\cite{MaherFischerGallyetal2017})
employ various convex relaxation schemes to obtain lower bounds on the
objective function value and use these bounds in divide-and-conquer
strategies to obtain {\it globally optimal solutions} \cite{ts05}.
We note that our MINLP approach bears similarities to the one presented in  \cite{horesh2016tech}, however unlike our model, the one presented in \cite{horesh2016tech} is computationally impractical.  

\newcommand{\R}{\mathbb{R}}
\newcommand{\B}{\{0,1\}}
\newcommand{\sm}{\setminus}
\newcommand{\etx}{expression tree}\newcommand{\et}{{\etx} }
\label{sec:problem}
We next give the details of our formulation.  The input to the
formulation consists of a rooted binary tree, a set of candidate
operators and an observed dataset.  For each observation $i\in I$, we
denote the value of the independent variables by $x^i\in\R^m$ and the
dependent variable values by $y^i\in\R$.  Let $D=\{1,\ldots,m\}$
denote the indices of the independent variables.  Let $T=(N,E)$ be the
input binary tree, let $s\in N$ denote root node and $L\subset N$
denote the leaf nodes.  Each node $n\in N\setminus\{s\}$ in the tree
has exactly one predecessor and each node $n\in N\setminus L$ has
exactly two successors.  Let $O=U\cup B$ denote the set of candidate
operators where $U$ contains the unary operators (such as squareroot)
and $B$ contains the binary operators (such as addition).  Our
formulation has two parts: The first part consists of constraints and
decision variables that construct the \et and the second part is used
to evaluate the difference between the estimated and actual dependent variable value of
each observation using the constructed \etx.

{\bf \ref{eq:G}.} The first part of the formulation chooses a subtree
of $T$ and assigns either an operator, a constant, or an input
variable to each node of the chosen subtree in a consistent way.  For
each $n\in N$, we have a decision variable $u_n\in\B$ to indicate if
the node is used (active) in the \etx.  
Decision
variables $z_{n,o}\in\B$ and $w_{n,d}\in\B$ denote whether the operator $o\in O$
 or the input variable
$d\in D$ are assigned
to node $n$ respectively. 
Finally we have a decision variable
$q_n\in\B$ if a constant value is assigned the node.  The constraint
\begin{equation}q_n +\sum_{d\in D} w_{n,d}+\sum_{o\in O} z_{n,o}= u_n, ~~ \forall n\in N,\label{eq:ops}\end{equation}
enforces that each active node is assigned an operator, a constant, or
one of the variables.  Let $n\in N\sm L$ be a non-leaf node with
successors $l,r\in N$ where $l$ has a smaller index than $r$.  Note
that if node $n$ is assigned a binary operator then both nodes $l,r$
have to be active in the tree; and if $n$ is assigned a unary operator
then exactly one of the nodes (say $l$) has to be active.
Consequently, we have constraints $u_r= \sum_{o\in B} z_{n,o}$ and
$u_l= \sum_{o\in B\cup U} z_{n,o}$.  Notice that these constraints
also enforce that if a node is inactive, or is assigned a constant or
is a variable, then its successor nodes cannot be active.  As the leaf
nodes cannot be assigned operators, we also set $ z_{n,o}=0$ for all
${o\in O}, l\in L$.  Finally, we define a continuous variable $c_n$
for $n\in N$ to denote the value of the constant when $q_n=1$.  It is
possible to show that any $c\in\R^{|N|}$ together with binary vectors
$q,w,z$, and $u$ (of appropriate dimensions) that satisfy the
constraints above define an \et and vice versa.

In addition to the basic model described above, we impose some additional constraints to improve computational performance (without compromising optimality).
For example we make sure that if a node is assigned a binary operator, then both its successor nodes cannot be assigned constants at the same time. Similarly the left successor of a unary operator cannot be a constant.
We also have several constraints to deal with symmetry as \etx s are inherently symmetric objects in the sense that the same function can be represented with different trees, for example by flipping two branches succeeding a commutative binary operator.
To avoid numerical problems, we also set bounds on the absolute value of the continuous variables.

{\bf \ref{eq:M}.  }  Using the first set of variables, the second part
of the formulation computes the value of the function (defined by the
\etx) for each observation.  More precisely, we define a variable
$v_{n,i}$ to denote the value of node $n\in N$ for observation $i\in
I$.  Value of a node clearly depends on the operator assigned to it
and the values of its successor nodes.  For example, if the addition
operator $a\in B$ is assigned to node $n\in N$ that has successor nodes
$r,l\in N$, then $v_{n,i}$ would be the sum of $v_{r,i}$ and
$v_{l,i}$.  Therefore we have a constraint:
\begin{equation}v_{n,i}=q_nc_n+\sum_{d\in D}w_{n,d}x^i_d+\sum_{o\in B}z_{n,o}f_o(v_{r,i},v_{l,i})+\sum_{o\in U}z_{n,o}f_o(v_{l,i})~~~~\forall i\in I,~n\in N,\label{eq:val}\end{equation}
where $f_o(\cdot)$ denotes the function on the argument(s) of the
operator $o\in O$.  Also note that $v_{n,i}\not=0$ only if $u_n=1$.
Leaf nodes cannot be assigned operators, therefore the last two
summations in \eqref{eq:val} are zero for these nodes.

{\bf \ref{eq:E}. }Recall that $s\in N$ denotes the root
node of the \et and therefore $v_{s,i}$ corresponds to the estimated
output variable for observation $i\in I$.  We define the error for
observation $i\in I$ as $(v_{s,i}-y^i)$ and relative error as
$(v_{s,i}/y^i-1)$.  
Using these expressions, we can define a discrepancy function such as $\mathcal{D}(v_{s},y)=\sum_{i\in
  I}(v_{s,i}-y^i)^2$.  
In our computational experiments, we control the relative error in our model by adding the constraint
$\sum_{i\in I}(v_{s,i}/y^i-1)^2\le \epsilon $ for a given $\epsilon > 0$.

{\bf \ref{eq:C}. } We can define a function to measure description
complexity of the model in various ways, for example
$\mathcal{C}(f_{cqwz})=\sum_{n\in N}\sum_{o\in O}c_{o}z_{n,o},$ where $c_o\in \R$ is
a complexity weight assigned to  operator $o\in O$ by the user.  In our
computational experiments, we use the expression $\sum_{n\in N}u_n$ to
minimize the total number of nodes in the \etx.

\section{Computational Experiments}
We report our computational experience on the discovery of physical
laws using real-world datasets. We solve the optimization problem
described in Section \ref{sec:problem} with the MINLP solver BARON
\cite{ts05} v17.8.9, using IBM ILOG CPLEX and IPOPT
\cite{ipopt} as subsolvers. We can only use operators that are handled
by BARON. Furthermore, when multiple globally optimal solution exists,
we do not have control over which one is returned. Experiments are
run on the cloud, so CPU speed cannot be precisely quantified. 

{\bf Exoplanet Data.}  The first set of experiments concerns the
discovery of Kepler's law on planetary motion  on a dataset taken from NASA
\cite{NASAExoplanetArchive} reporting information on planetary
systems. The dataset consists of tuples of the form (planet, star,
$\tau, M, m, d$) where $\tau$ is the orbital period of the planet, $M$
the mass of the star, $m$ the mass of the planet, $d$ the major
semi-axis of the orbit. Planet features are normalized to Earth's,
star features to Sol's. $\tau$ is further normalized since its range
is very large. We build three test problems from this dataset,
including the systems listed in brackets: EP1 (Sol and Trappist), EP2
(HD 219134 and Kepler), EP3 (GJ 667C, HD 147018, HD 154857, HD 159243
and HD 159868).
\begin{table}[tb]
  \centering
  \caption{Results for the exoplanet dataset. Time limit is 6
    hours.  Grey cells exceed the time limit. 
    $c$ represents constants   (we   do not report its value for brevity).}  \small
  \begin{tabular}{|l|*{6}{c|}}
    \hline
    & \multicolumn{6}{c|}{Maximum relative error} \\
    \cline{2-7}
    Dataset & 2\% & 5\% & 10\% & 20\% & 30\% & 50\% \\
    \hline
    EP1 & $\sqrt[3]{c\tau^2M}$ & $\sqrt[3]{c\tau^2M}$ & $\sqrt[3]{\tau^2}(M+c)$ & $\sqrt[3]{\tau}(\sqrt{\tau} + M)$ & $\sqrt{c\tau^2}$ & $\sqrt{\tau M}$ \\
    EP2 & $\sqrt[3]{c\tau^2M}$ & $c\sqrt[3]{\tau^2}$ & $c\sqrt[3]{\tau^2}$ & $\sqrt[3]{c\tau^2}$ & $\sqrt{\tau}$  & $\sqrt{\tau}$ \\
    EP3 & \cellcolor{gray!25} $\sqrt[3]{c\tau^2M}$ & $\sqrt[3]{c \tau^2M}$ & $\sqrt{\tau M} + \tau$ & \cellcolor{gray!25} $\sqrt[3]{c\tau^2M}$ & $\sqrt{c\tau} + c$ &  $\sqrt{\tau}$ \\
    \hline
  \end{tabular}\label{tab:ep}
\end{table}

We consider the set of operators $+,*,\sqrt{},\sqrt[3]{}$, and
full binary expression trees with depth at most 3 (the root
has depth 0) in which each operator can appear at most 3 times. We aim
to predict $d$ in terms of the other input variables. Results are
reported in Table \ref{tab:ep}, for different upper bounds applied to
the relative model error. For small model errors, we always find a
refined expression of Kepler's third law: $d =
\sqrt[3]{c\tau^2M}$. Given the data, the more comprehensive formula
would be $d = \sqrt[3]{c\tau^2(M+m)}$, but the dependency on $m$ is
not picked up because $m$ is negligible compared to $M$. As model
error upper bound increases, we find simpler formulas that do not
accurately reflect Kepler's third law.

Most of the formulas returned are certified globally optimal in a
relatively short period of time: on average, 50 minutes for the
instances solved to global optimality (two instances hit the time
limit and thus are not certified globally optimal within the time limit), standard deviation 75.  The number of nodes explored by the
Branch-and-Bound algorithm varies greatly: from 147 for the simplest
formula $\sqrt{\tau}$, to over 600k for one instance that is not
solved to optimality within the allotted runtime. The geometric mean
is 19917. By contrast, the number of binary expression trees of depth
$3$ with $4$ candidate operations, $3$ input variables, and numerical
constants is $\approx 2 \cdot 10^7$ without accounting for
symmetry. We remark that the number of nodes explored by
Branch-and-Bound does not correspond to the number of candidate
expression trees that have been examined: it is merely a measure of
difficulty of the search.

{\bf Pendulum Data.}  The second set of experiments concerns ten
pendulums with different link lengths, and timestamps for the time at
which each pendulum crosses the midpoint of its arc. The data is
obtained from a high-resolution video recording of the pendulums,
yielding hundreds of timestamps. From the raw data, we randomly
extract ten tuples (one for each link length) of the form ($\ell, i,
t_i, j, t_j$), where $\ell$ is the link length, $i$ and $j$ are
integers, $t_i$ and $t_j$ are the timestamps for the $i$-th and $j$-th
crossing of the midpoint, and $j > i$. We aim to predict $t_j$.
\begin{table}[tb]
  \centering
  \caption{Results for the pendulum dataset. Time limit is 6 hours.
    All formulas are certified optimal.}
    \small
  \begin{tabular}{|l|*{4}{c|}}
    \hline
    & \multicolumn{4}{c|}{Maximum relative error} \\
    \cline{2-5}
    Dataset & 0.5\% & 1\% & 2\% & 5\% \\
    \hline
    DS1 & $j(\ell - c) -\ell$  & $j\sqrt{\ell} + c$  & $j\sqrt{\ell}$ & $cj$  \\
    DS2 & $j\sqrt{\ell}$  & $j\sqrt{\ell}$  & $cj$ & $cj$  \\
    DS3 & $j\sqrt{\ell}$  & $j\sqrt{\ell}$  & $\ell - cj$ & $cj$  \\
    DS4 & $j(\ell -c) -c$  & $j\sqrt{\ell}$  & $cj$ & $cj$  \\
    \hline
  \end{tabular}\label{tab:ds}
\end{table}

The period of the pendulum is given by $\tau =
2\pi\sqrt{\ell/g}$. Hence, we aim to obtain the equation $t_j = \pi
j\sqrt{\ell/g}$. The experimental setup is similar to the exoplanet
dataset, but we add ``$-$'' to the list of operators. 
Table \ref{tab:ds} shows that many experiments
return the correct formula except the multiplicative constant, since
$\pi/\sqrt{g} \approx 1.002$ and its discrepancy from $1$ is too small
to be significant, up to experimental error. With error bound $0.5\%$,
we observe overfitting on DS1 and DS4: the simpler formula
$j\sqrt{\ell}$ does not satisfy the error bound due to experimental
error.  Because the link lengths of the ten pendulums are close to each
other (ranging from $0.28m$ to $0.307m$), for larger error bounds a
simpler expression of the form $cj$ suffices. The average runtime is 2
minutes, standard deviation 2.3. The number of nodes ranges from 57 to
8134, geometric mean 299. Despite the presence of redundant variables,
the algorithm quickly identifies the simplest formula explaining the
data.

In this extended abstract we focused on small datasets and did not
address scalability, which is a known limitation of
Branch-and-Bound-based methods for MINLP. This is left for future research.

\newpage

{\small
  \bibliographystyle{plainurl}
  \bibliography{refs}

\begin{thebibliography}{10}

\bibitem{augusto2000symbolic}
Douglas~Adriano Augusto and Helio~JC Barbosa.
\newblock Symbolic regression via genetic programming.
\newblock In {\em Proceedings of the Sixth Brazilian Symposium on Neural
  Networks}, pages 173--178. IEEE, 2000.

\bibitem{banzhaf1998genetic}
Wolfgang Banzhaf, Peter Nordin, Robert~E Keller, and Frank~D Francone.
\newblock {\em Genetic programming: an introduction}, volume~1.
\newblock Morgan Kaufmann, San Francisco, 1998.

\bibitem{BeLeLiMaWa08}
Pietro Belotti, Jon Lee, Leo Liberti, Fran{\c{c}}ois Margot, and Andreas
  W{\"a}chter.
\newblock Branching and bounds tightening techniques for non-convex {MINLP}.
\newblock {\em Optimization Methods and Software}, 24(4-5):597--634, 2008.

\bibitem{brunton2016discovering}
Steven~L Brunton, Joshua~L Proctor, and J~Nathan Kutz.
\newblock Discovering governing equations from data by sparse identification of
  nonlinear dynamical systems.
\newblock {\em Proceedings of the National Academy of Sciences},
  113(15):3932--3937, 2016.

\bibitem{Luo2017}
Zonglin~Jiang Changtong~Luo, Chen~Chen.
\newblock A divide and conquer method for symbolic regression.
\newblock {\em arXiv:1705.08061v2}, 2017.

\bibitem{connor1977scaling}
Jack~W Connor and John~B Taylor.
\newblock Scaling laws for plasma confinement.
\newblock {\em Nuclear Fusion}, 17(5):1047, 1977.

\bibitem{davidson1999method}
James~W Davidson, Dragan~A Savic, and Godfrey~A Walters.
\newblock Method for the identification of explicit polynomial formulae for the
  friction in turbulent pipe flow.
\newblock {\em Journal of Hydroinformatics}, 1(2):115--126, 1999.

\bibitem{davidson2003symbolic}
James~W Davidson, Dragan~A Savic, and Godfrey~A Walters.
\newblock Symbolic and numerical regression: experiments and applications.
\newblock {\em Information Sciences}, 150(1):95--117, 2003.

\bibitem{horesh2016tech}
Lior Horesh, Leo Liberti, and Haim Avron.
\newblock Globally optimal mixed integer non-linear programming ({MINLP})
  formulation for symbolic regression.
\newblock Technical Report 219095, IBM, 2016.

\bibitem{dgp}
Dario Izzo, Francesco Biscani, and Alessio Mereta.
\newblock Differentiable genetic programming.
\newblock In J.~McDermott et~al., editor, {\em EuroGP 2017, LNCS 10196}, pages
  35--51, 2017.

\bibitem{koza}
John~R. Koza.
\newblock {\em Genetic Programming: On the Programming of Computers by Means of
  Natural Selection}.
\newblock MIT Press, Cambridge, 1992.

\bibitem{kroll2017workflow}
Paul Kroll, Alexandra Hofer, Ines~V Stelzer, and Christoph Herwig.
\newblock Workflow to set up substantial target-oriented mechanistic process
  models in bioprocess engineering.
\newblock {\em Process Biochemistry}, 2017.

\bibitem{langley1981data}
Pat Langley.
\newblock Data-driven discovery of physical laws.
\newblock {\em Cognitive Science}, 5(1):31--54, 1981.

\bibitem{langley1987}
Pat Langley, Gary~Lee Bradshaw, and Herbert~A Simon.
\newblock Heuristics for empirical discovery.
\newblock In L.~Bolc, editor, {\em Computational models of learning}. Springer,
  Berlin, 1987.

\bibitem{MaherFischerGallyetal2017}
Stephen~J. Maher, Tobias Fischer, Tristan Gally, Gerald Gamrath, Ambros
  Gleixner, Robert~Lion Gottwald, Gregor Hendel, Thorsten Koch, Marco~E.
  L{\"u}bbecke, Matthias Miltenberger, Benjamin M{\"u}ller, Marc~E. Pfetsch,
  Christian Puchert, Daniel Rehfeldt, Sebastian Schenker, Robert Schwarz,
  Felipe Serrano, Yuji Shinano, Dieter Weninger, Jonas~T. Witt, and Jakob
  Witzig.
\newblock The {SCIP} optimization suite 4.0.
\newblock Technical Report 17-12, ZIB, Takustr.7, 14195 Berlin, 2017.

\bibitem{ffx}
T.~McConaghy.
\newblock Ffx: Fast, scalable, deterministic symbolic regression technology.
\newblock In {\em Genetic Programming Theory and Practice X}, pages 235--260.
  Springer, New York, 2011.

\bibitem{NASAExoplanetArchive}
NASA.
\newblock {NASA} exoplanet archive, 2017.
\newblock URL: \url{https://exoplanetarchive.ipac.caltech.edu/docs/intro.html}.

\bibitem{ge}
M.~O’Neill and C.~Ryan.
\newblock Grammatical evolution.
\newblock {\em IEEE Trans. Evol. Comput.}, 5(4):349--358, 2001.

\bibitem{schmidt2009distil}
Michael Schmidt and Hod Lipson.
\newblock Distilling free-form natural laws from experimental data.
\newblock {\em Science}, 324(3):81--85, 2009.

\bibitem{schmidt2010symbolic}
Michael Schmidt and Hod Lipson.
\newblock Symbolic regression of implicit equations.
\newblock In {\em Genetic Programming Theory and Practice VII}, pages 73--85.
  Springer, Berlin, 2010.

\bibitem{ts05}
Mohit Tawarmalani and Nikolaos~V Sahinidis.
\newblock A polyhedral branch-and-cut approach to global optimization.
\newblock {\em Mathematical Programming}, 103(2):225--249, 2005.

\bibitem{todorovski1997declarative}
Ljupco Todorovski and Saso Dzeroski.
\newblock Declarative bias in equation discovery.
\newblock In {\em Proceedings of the Fourteenth International Conference on
  Machine Learning}, pages 376--384, 1997.

\bibitem{topchy}
A.~Topchy and W.~F. Punch.
\newblock Faster genetic programming based on local gradient search of numeric
  leaf values.
\newblock In {\em Proceedings of the Genetic and Evolutionary Computation
  Conference}, pages 155--162, 2011.

\bibitem{ipopt}
Andreas W\"achter and Larry~T Biegler.
\newblock On the implementation of a primal-dual interior point filter line
  search algorithm for large-scale nonlinear programming.
\newblock {\em Mathematical Programming}, 106(1):25--57, 2006.

\bibitem{willis1997genetic}
Mark~J Willis, Hugo~G Hiden, Peter Marenbach, Ben McKay, and Gary~A Montague.
\newblock Genetic programming: An introduction and survey of applications.
\newblock In {\em Genetic Algorithms in Engineering Systems: Innovations and
  Applications, 1997. GALESIA 97. Second International Conference On (Conf.
  Publ. No. 446)}, pages 314--319. IET, 1997.

\end{thebibliography}
}

\end{document}